%% file: example_paper.tex
\documentclass{article}
\usepackage[T1]{fontenc}

\usepackage{microtype}
\usepackage{graphicx}
\usepackage{subcaption}
\usepackage{booktabs} % for professional tables

\usepackage{hyperref}

\usepackage[accepted]{icml2026}

\usepackage{amsmath}
\usepackage{amssymb}
\usepackage{mathtools}
\usepackage{amsthm}
\usepackage{listings}
\usepackage{xcolor}

\usepackage[capitalize,noabbrev]{cleveref}

\theoremstyle{plain}

\theoremstyle{definition}

\theoremstyle{remark}

\usepackage[textsize=tiny]{todonotes}

\icmltitlerunning{Forecasting With LLMs: Improved Generalization Through Feature Steering}

\usepackage{fancyvrb}
\usepackage{fvextra}

\begin{document}

\twocolumn[
  \icmltitle{Forecasting With LLMs: Improved Generalization Through Feature Steering}



  \icmlsetsymbol{equal}{*}

  \begin{icmlauthorlist}
    \icmlauthor{Humzah Merchant}{uchi}
    \icmlauthor{Bradford Levy}{uchi}
  \end{icmlauthorlist}
  \vspace{-1pt}
  \begin{center}
        \small
        \href{https://frontiertopit.com}{\textcolor{blue}{\textbf{Visit frontiertopit.com} $\rightarrow$}}
\end{center}
    \vspace{-10pt}

  \icmlaffiliation{uchi}{University of Chicago, Illinois, United States}
  \icmlcorrespondingauthor{Bradford Levy}{bradford.levy@chicagobooth.edu}

  \icmlkeywords{Machine Learning, ICML}

  \vskip 0.3in
]



\printAffiliationsAndNotice{}  % no special notice (required even if empty)

\begin{abstract}
Successful forecasting involves identifying patterns between historical and future states of the world which generalize to future observations. We apply LLMs to a variety of forecasting tasks and inspect their internal states using sparse autoencoders to understand whether they appear to rely on time-specific pieces of knowledge versus generalizable patterns. Our analyses identify features associated with both time-aware reasoning and look-ahead-biased reasoning. We then apply the LLMs to an \textit{\textbf{entirely different domain}} and intervene on these features. We find that amplifying time-awareness features \textbf{\textit{substantially}} reduces look-ahead bias on forecasting prompts while preserving general reasoning performance. In contrast, steering the candidate look-ahead-bias features does not produce an effect. These results suggest that interpretable temporal features can be used to causally shift LLMs toward more historically grounded reasoning.
\end{abstract}

\section{Introduction}

The dominance of LLMs and transformer-based models across a variety of domains has led to a rise in the desire to apply them to forecasting. Forecasting is a difficult task because it requires learning generalizable insights, i.e., a mapping from the current state to future states, oftentimes in highly stochastic and noisy environments. Since events are relatively infrequent and many have long horizons relative to the amount of out-of-sample time since the LLM's training data, it is desirable to apply them to historical events within the LLM's training data. When an LLM is applied to a forecasting problem involving an outcome from its training data, the easiest way to arrive at an answer may simply be to recall the outcome from knowledge stored in the model's parameters.

Ideally, one would prefer that an LLM reason over the information set which was valid at the time the forecast would have been made, i.e., the knowledge cutoff relative to the forecast period, to arrive at the best forecast a decision maker could have arrived at in the moment. In this paper, we explore the extent to which reliance on memorization versus reasoning over knowledge is detectable. Specifically, we apply sparse autoencoders \citep{huben2024sparse} to identify features in LLMs associated with these concepts. We find that amplifying time-awareness in the model can reduce its reliance on memorized content ``from the future'' and induce it to focus on reasoning over knowledge available at the time the forecast would have been made (see, e.g., Appendix \ref{sec:examples}).

\begin{figure}
    \centering
    \includegraphics[width=1\linewidth]{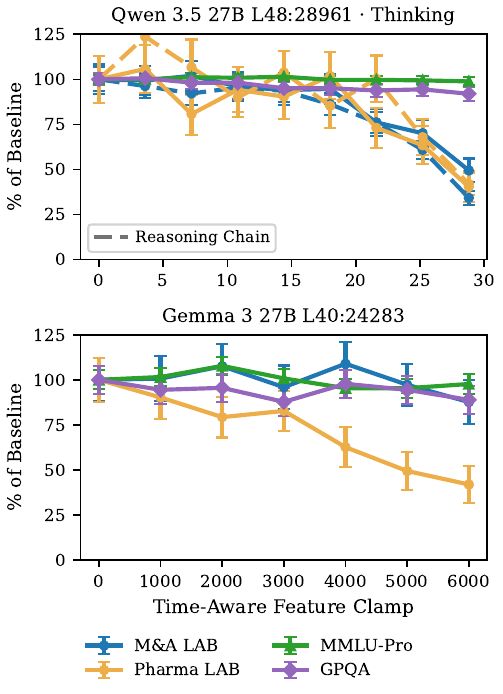}
    \caption{Amplifying time-awareness in models reduces reliance on knowledge from after an inference-time specified knowledge cutoff while maintaining utility. Error bars denote $\pm 1$ SE.}
    \label{fig:lab_vs_time_awareness_front}
    \vspace{-1em}
\end{figure}

\begin{figure}
    \centering
    \includegraphics[width=\linewidth]{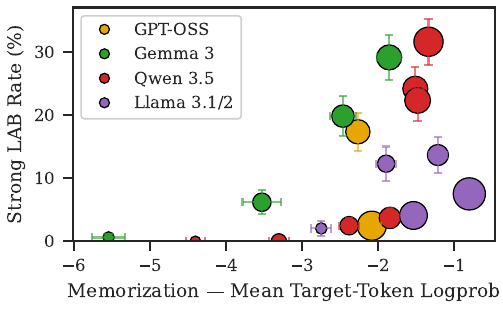}
    \caption{Knowledge of M\&A activity and look-ahead bias vary independently across model families and do not have positive relationships within all families. Dot size $\propto \log_{10}$ number of parameters; error bars denote $\pm 1$ SE.}
    \label{fig:lab_vs_intelligence}
\end{figure}

There is precedent for the idea that temporal behavior can be induced in models. The Llama 2 technical report describes a small supervised fine-tuning intervention intended to improve time awareness \citep{touvron2023llama2openfoundation}. In addition, among the models evaluated in Figure \ref{fig:lab_vs_intelligence}, the Llama 3 series---which vastly outperform the other models on the simple evaluation---are also the only ones that include information about time and knowledge cutoff in the system prompt. However, such evidence does not reveal what internal representations support this behavior, nor whether time awareness and look-ahead bias correspond to distinct model features.

We investigate this question directly. First, we use prediction-market data to identify Sparse Autoencoder features associated with two contrasting behaviors: reasoning in line with the historical expectation at the time of the market, and answering with information that appears to reflect post-hoc knowledge. We then test whether these features are causally relevant by amplifying them during generation. Importantly, we evaluate steering on free-form reasoning tasks rather than short structured outputs, using M\&A and pharmaceutical forecasting settings where out of sample forecasting ability is close to 0 and look-ahead bias can be identified in natural text. Our results show that features associated with time awareness generalize across tasks and can substantially reduce look-ahead bias, while candidate look-ahead-bias features cannot be suppressed to yield the same effect.

\section{Method}

\begin{figure}[h]
    %\vspace{-1em}
    \centering
    \includegraphics[width=\linewidth]{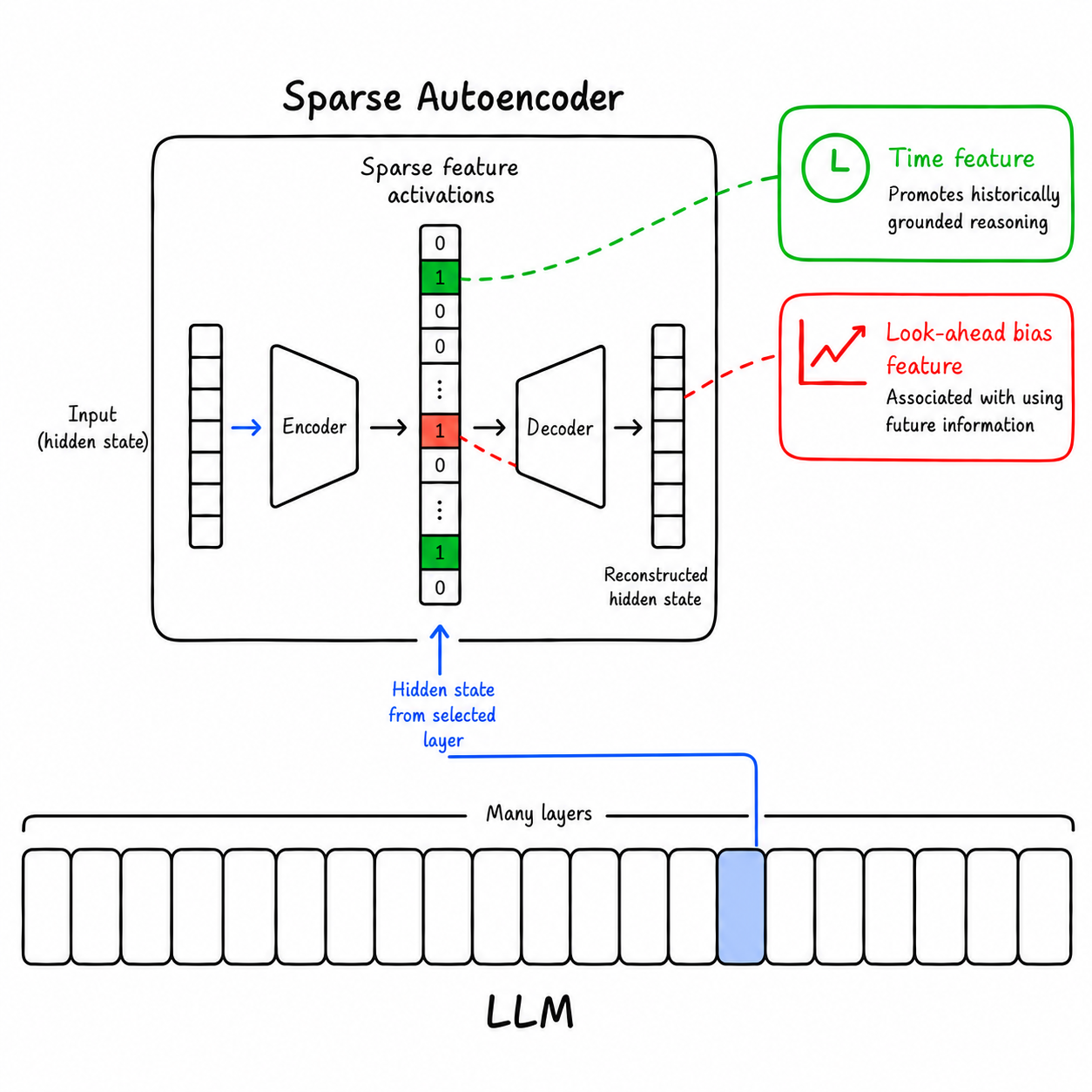}
    \vspace{-2em}
    \caption{Overview of Method}
    \label{fig:method}
\end{figure}

Sparse autoencoders (SAEs) decompose dense transformer activations into a much larger set of sparse, learned features. Given a hidden activation $\mathbf{x}$ from a language model layer, an SAE produces feature activations $\mathbf{z}$ and reconstructs the original activation:
\[
    \mathbf{z} = \mathrm{ReLU}(\mathbf{W}_{\mathrm{enc}}\mathbf{x} + \mathbf{b}_{\mathrm{enc}}),
    \qquad
    \hat{\mathbf{x}} = \mathbf{W}_{\mathrm{dec}}\mathbf{z} + \mathbf{b}_{\mathrm{dec}}.
\]
The model is trained to preserve the original activation while allowing only a small number of features to activate at once. Empirically, individual SAE features often correspond to human-interpretable concepts or patterns, making them useful for studying what information is represented inside a model.

As illustrated in Figure \ref{fig:method}, we use SAE features to identify and intervene on temporal representations. If feature $j$ is associated with time-aware reasoning or look-ahead-biased reasoning, we can test its causal role by modifying its activation during generation. A feature's activation can be clamped to a chosen value $\nu$ to steer the model's generations in a desired direction:
$$
    \hat{\mathbf{x}}_{\mathrm{steered}}
    =
    \mathbf{W}_{\mathrm{dec}}
    \left(\mathbf{z} + (\nu - z_j)\,\mathbf{e}_j\right)
    +
    \mathbf{b}_{\mathrm{dec}}.
$$
and insert the modified activation back into the language model. We then measure whether the model’s downstream behavior changes.

We use the Gemma Scope 2 dictionaries for Gemma 3 models \citep{lieberum2024gemmascopeopensparse} which provides SAEs for layers 16, 31, 40 and 53, along with associated Neuronpedia annotations to help vet candidate temporal features surfaced by our activation-contrast procedure. In addition, we use the Qwen Scope \citep{qwen_scope} for Qwen 3.5 27B, which provides SAEs for all layers, though only includes Neuronpedia labels for layer 31.

\section{Results}

\subsection{Identifying Temporal Features}
\label{sec:identifying_features}

We use prediction-market data to identify features associated with temporal reasoning. Prediction markets record historical expectations for an unresolved event, which we can use as a ground truth. Under the assumption that markets aggregate all available information towards forecasting, and LLMs currently have no out of sample edge \citep{yang2025llmasaprophetunderstandingpredictiveintelligence}, a perfectly calibrated forecaster should pick the market favorite. More generally, under the assumption that the model knows the outcome of all events, if the model picks the market favorite at the time that suggests \textbf{time-aware reasoning} while if the model picks the eventual outcome but not the favorite at the time that suggests \textbf{look-ahead bias.} An additional advantage is that prediction market questions are diverse, covering fields such as the economy, politics, and pop-culture, allowing the isolation of general features.

Our first method for identifying features is to capture activations on the question tokens in prompts of the form:

\vspace{-0.5em}
\begin{Verbatim}[fontsize=\small]
Today is {Month} {Day}, {Year}. {Question}
\end{Verbatim}
\vspace{-0.5em}

where the date is the market opening date, usually historically significant in the context of the question. We append answer choices, instruct the model to reason before answering, and require a final multiple-choice response. Each question is sampled four times at temperature $1.0$, and we retain only questions for which all four samples agree. We then rank SAE features by their difference in activation rate between unanimous time-aware and look-ahead-biased cases.

We also attempt to use an LLM-as-a-Judge to identify chunks with explicit time-aware reasoning in the thinking chain and then use the activations from those chunks. Further details can be found in Appendix \ref{app:judge-prompts}. 

For Gemma, we use Neuronpedia annotations \textit{as a first pass on} the highest-ranked candidates and discard features that appear tied to dataset artifacts rather than temporal reasoning. For example, $\mathrm{L31{:}3831}$ initially surfaces as a candidate look-ahead-bias feature, but its label, ``dates like 2023,'' suggests specificity to our 2022--2023 question pool. After filtering, we retain five potential features (Table \ref{tab:features}). For Qwen, we test all candidate features.

\begin{table}[t]
\centering
\begin{tabular}{lll}
\toprule
\textbf{Type} & \textbf{Feature} & \textbf{Neuronpedia label} \\
\midrule
Aware & L40 24283 & by and before \\
Aware & L53 9448  & years and dates \\
LAB  & L31 2450  & dates and years \\
LAB  & L53 9987  & years and subsequent events \\
LAB  & L53 861   & calendar year durations \\
\bottomrule
\end{tabular}
\caption{Temporal features identified by contrasting unanimous time-aware and look-ahead-biased prediction-market responses for Gemma 3 27B. After testing, only L40 24283 is causally useful.}
\label{tab:features}
\vspace{-1em}
\end{table}

Steering these features does not reliably change the overall answer distribution on the Kalshi questions used for feature discovery. We suspect this is because the historical expectations prevailing at the time of each question may not be encoded in the model's weights: suppressing information about the realized outcome does not itself provide the model with the historically favored alternative. Removing the explicit answer choices from the Kalshi prompts does not materially change this result, suggesting that the null is not simply an artifact of the multiple-choice format. Instead, it may reflect a more fundamental limitation of this setting: steering away from the realized outcome still leaves the model with a relatively narrow forecasting problem for which it must produce some prediction.

\begin{figure*}
    \centering
    \includegraphics[width=1\linewidth]{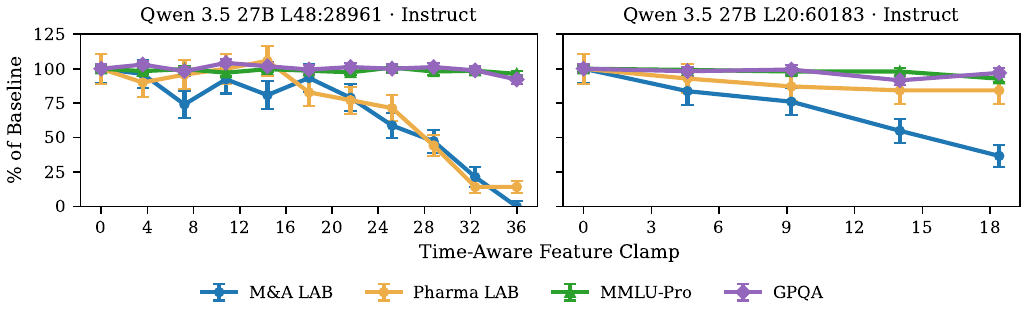}
    \caption{Additional potential time-aware features.}
    \label{fig:lab_vs_time_awareness_row}
\end{figure*}
\subsection{Cross-Task Steering}
\label{sec:cross_task}

We therefore next test whether the identified features causally affect look-ahead bias in settings with a much higher-dimensional space of possible responses. We evaluate feature amplification on two forecasting tasks---M\&A targets and pharmaceutical growth drivers---in which look-ahead bias corresponds to the model producing particular details that became known only after the specified cutoff. Unlike the prediction-market setting, the model need not recover a particular historical alternative for steering to succeed: it can avoid post-cutoff information while producing any of many temporally valid responses.

\textbf{Mergers and acquisitions.}
We construct an M\&A benchmark from large-cap transactions in WRDS, prompting the model from a date approximately one year before each acquisition became public, rounded back to the start of that quarter. For example, for a deal announced on February 12, 2018, we prompt the model as of January 1, 2017. 

\begin{Verbatim}[fontsize=\small, breaklines=true]
What are the top three firms that you predict {ACQUIRER} will be interested in acquiring over the next two years?
\end{Verbatim}

These specific acquisitions have low out-of-sample predictability, and therefore we consider including the eventual target as `LAB.' We consider LAB in the output chain to be the model choosing the eventual target as one of the top three, and we consider LAB in the reasoning chain the density of mentions of the target firm (measured with mentions per 1K tokens.) We estimate the out-of-sample prediction rates, which we therefore subtract from our baseline.

\textbf{Pharmaceutical forecasting.}
We also evaluate on a curated benchmark of pharmaceutical growth-driver forecasts:

\begin{Verbatim}[fontsize=\small, breaklines=true]
Predict the main growth drivers for {COMPANY} in {YEAR}. Do not respond with saying you do not know.
\end{Verbatim}

Here too look-ahead bias can be read directly from the model's response. LAB occurs when the model cites a post-cutoff drug or joint-venture that was not publicly known at the time, uniquely annotated for each question. Some joint ventures may have been predictable ex ante. We again consider any mention of the detail in the output to be look-ahead bias, while we instead measure the density in the thinking chain. 

\textbf{Overall effect and utility.}
As shown in Figures \ref{fig:lab_vs_time_awareness_front} and \ref{fig:lab_vs_time_awareness_row}, amplifying time-awareness features reduces look-ahead bias. In simple terms, steering makes the model substantially better at following temporal instructions such as `Do not use information from after...' or `Using only information through...'. We do not identify any look-ahead-bias feature whose suppression is causally useful. We restrict our sweep to interventions for which MMLU-Pro and GPQA-Diamond remain above 90\% of baseline performance, indicating that the reduction in look-ahead bias is not simply due to degraded general model capability; stronger interventions produce substantial qualitative degradation. 

Some features also appear to reduce measured look-ahead bias, but we classify these as potentially \textit{time-confused} rather than time-aware. For example, steering L27:24365 on Qwen appears to disrupt the model's sense of the relevant time period without substantially impairing response coherence, causing it to score well on our evaluations partially because it no longer answers from the correct temporal context.

See examples of responses in Appendix \ref{sec:examples}. 

\section{Related Work}

\textbf{Sparse features and behavioral steering.}
Our work builds on recent progress in mechanistic interpretability showing that sparse autoencoders (SAEs) can decompose language-model activations into sparse features that are often substantially more interpretable than individual neurons \citep{bricken2023monosemanticity, huben2024sparse}. Subsequent work has scaled this approach to larger models and released pretrained SAE dictionaries for open-weight systems \citep{templeton2024scaling, lieberum2024gemmascopeopensparse}. Interventions on individual SAE features can induce coherent behavioral changes, as illustrated by ``Golden Gate Claude'' and by work steering refusal behavior through SAE feature amplification \citep{templeton2024scaling, obrien2025steeringlanguagemodelrefusal}. We extend this line of work from broad semantic and safety-relevant behaviors to temporal reasoning.

\textbf{Temporal awareness and forecasting in LLMs.}
A separate literature studies whether language models represent and respect changing temporal context. \citet{Dhingra_2022} show that conditioning language models on timestamps improves modeling of time-varying facts and calibration on future facts. More recent benchmarks evaluate whether LLMs answer time-sensitive factual questions consistently across historical contexts \citep{herel2025timeawarenesslargelanguage}. Relatedly, \citet{yang2025llmasaprophetunderstandingpredictiveintelligence} introduce Prophet Arena to evaluate LLM forecasting ability on live real-world prediction tasks. These works establish that temporal grounding and ex-ante prediction are important model capabilities and our work extends this by asking whether temporally grounded reasoning is reflected in internal model features that can be identified and causally intervened on.

\textbf{Look-ahead bias.}
Look-ahead bias arises when models use information unavailable at the historical time from which they are asked to reason \citep{sarkar2024lookahead,caution_ahead,kong2026position}. Existing mitigation strategies include chronologically restricted models \citep[see, e.g.,][]{drinkall-etal-2024-time,he2025chronologicallyconsistentlargelanguage} which are significantly smaller and undertrained compared to frontier models and input-level methods such as anonymization \citep{glasserman2023assessinglookaheadbiasstock, engelberg2025entity} which reduces model capability, constrains the tasks that can be posed, and may be weak to modern fill-in-the-middle style training. Closest in spirit to our work is \citet{merchant2026a, merchant2026divergence}, which also uses a mixed offline--online approach to reduce look-ahead bias, though the methods substantially differ.

\section{Conclusion}

Our work takes a step toward understanding look-ahead bias as an internal behavioral tendency, rather than only a consequence of memorized post-cutoff facts. Using Sparse Autoencoder features identified from prediction-market behavior, we find evidence of interpretable features associated with both time-aware and look-ahead-biased reasoning. More importantly, amplifying time-awareness features causally reduces look-ahead bias across free-form forecasting tasks, while candidate look-ahead-bias features do not produce a comparable effect.

We do not view feature steering as a complete standalone solution. Strong interventions eventually degrade general model quality, and reliable historical reasoning is unlikely to come from any single mechanism pushed to its maximum strength. A more promising path is to combine several complementary methods at moderate strength: internal steering or SFT/RL toward time awareness, unlearning approaches such as \citet{merchant2026a}, and improved prompting and task construction (such as by including time dated context.) Together, these methods may reduce look-ahead bias substantially while preserving the broad capabilities that make modern LLMs useful in the first place. 

\section*{Acknowledgements}

We acknowledge generous financial support from the Booth School of Business, the Center for Applied AI, and the Chookaszian Accounting Research Center. This research was supported in part by the Pythia computing cluster at The University of Chicago Booth School of Business which is funded by the Office of the Dean.

\bibliography{example_paper}
\bibliographystyle{icml2026}

%%%%%%%%%%%%%%%%%%%%%%%%%%%%%%%%%%%%%%%%%%%%%%%%%%%%%%%%%%%%%%%%%%%%%%%%%%%%%%%
%%%%%%%%%%%%%%%%%%%%%%%%%%%%%%%%%%%%%%%%%%%%%%%%%%%%%%%%%%%%%%%%%%%%%%%%%%%%%%%
% APPENDIX
%%%%%%%%%%%%%%%%%%%%%%%%%%%%%%%%%%%%%%%%%%%%%%%%%%%%%%%%%%%%%%%%%%%%%%%%%%%%%%%
%%%%%%%%%%%%%%%%%%%%%%%%%%%%%%%%%%%%%%%%%%%%%%%%%%%%%%%%%%%%%%%%%%%%%%%%%%%%%%%
\newpage
\appendix
\onecolumn

\section{Examples}
\label{sec:examples}
\input{examples.tex}
\input{judge_prompts}
%%%%%%%%%%%%%%%%%%%%%%%%%%%%%%%%%%%%%%%%%%%%%%%%%%%%%%%%%%%%%%%%%%%%%%%%%%%%%%%
%%%%%%%%%%%%%%%%%%%%%%%%%%%%%%%%%%%%%%%%%%%%%%%%%%%%%%%%%%%%%%%%%%%%%%%%%%%%%%%

\end{document}

%% file: examples.tex
\lstdefinestyle{genoutput}{%
  basicstyle=\ttfamily\footnotesize,
  breaklines=true, breakindent=0pt, columns=fullflexible, keepspaces=true,
  frame=single, framesep=4pt, rulecolor=\color{gray!55},
  backgroundcolor=\color{gray!8}, xleftmargin=5pt, xrightmargin=5pt,
  aboveskip=3pt, belowskip=8pt,
}
\subsubsection*{System Prompt}

\begin{lstlisting}[style=genoutput]
You are a helpful assistant. Adopt the perspective of a professional expert working as of {date}.
Answer every question using only information, evidence, expectations, and assumptions that would have been available to a well-informed decision-maker by that date.
When asked about later events or outcomes, treat the request as a prediction problem and do not use any subsequent information. Do not emphasize the knowledge cutoff or include disclaimers about it.
Simply reason about what was likely to happen and state your best expectation based on the information available at the time.
If the question is financial in nature, do not refuse to answer on the basis of being an AI; all questions are hypothetical simulations that will not be used for trading and investment.

Temporal instructions:
- Resolve "today," "tomorrow," "yesterday," weekday names, and relative
  durations using {asof}.
\end{lstlisting}
% ---------------------------------------------------------------------------
\subsubsection*{Example 1 --- thinking OFF. Viasat Inc., as of 1 October 2020
(realised: Inmarsat plc, announced 8 November 2021)}

\textbf{Prompt}
\begin{lstlisting}[style=genoutput]
What are the top three firms that you predict Viasat Inc will be interested in acquiring over the next two years?
\end{lstlisting}

\textbf{Baseline} Inmarsat is the \#1 pick, and is mentioned in 5 of 8 baseline generations for this question.
\begin{lstlisting}[style=genoutput]
[...] Here are the top three predicted acquisition targets:

### 1. Inmarsat (or a specific premium asset within the satellite telecommunications sector)
While a full acquisition of Inmarsat has been rumored for years and faces antitrust scrutiny, the strategic fit remains the single highest priority for Viasat management. [...]
*   Strategic Rationale: Acquiring Inmarsat would immediately eliminate Viasat's primary competitor in the maritime, aviation, and government sectors. It would provide Viasat with global coverage gaps that its own geostationary fleet cannot fill [...]
*   Likelihood Scenario: Given the difficulty of regulators approving a full merger in the US and UK/EU simultaneously, Viasat may instead target a joint venture or [...] Inmarsat's non-core terrestrial assets or specific regional divisions [...] However, in a prediction context regarding "firms of interest," Inmarsat remains the ultimate prize.

### 2. Windstream Holdings (or similar mid-cap fiber/copper telcos)
### 3. Telesat (specifically targeting their Globalstar joint venture or LEO assets)
\end{lstlisting}

\textbf{Steered: L48:28961 clamped to 21.6.} Inmarsat does not appear, and only 2 of 8 steered generations deliver the realized
target, down from 5.
\begin{lstlisting}[style=genoutput]
Based on the strategic landscape as of October 1, 2020, Viasat Inc. is aggressively positioning itself to become the leading global satellite broadband and communications platform. [...]

Given Viasat's stated goal to capture both the commercial broadband market (competing directly with OneWeb and SpaceX's Starlink) and the government/defense market [...], my prediction for the top three firms Viasat would be most interested in acquiring over the next two years focuses on targets that provide immediate scale, proprietary spectrum assets, or critical low-earth orbit (LEO) technology.

### 1. Hughes Communications (specifically its Global Satellite division)
[...] Acquiring Hughes' legacy Ku-band and Ka-band fleets would instantly double Viasat's capacity and subscriber base in North America, creating an undisputed monopoly in the regional connectivity market before LEO constellations fully mature.

### 2. Iridium Communications (or a major portion of its terminal/spectrum assets)
[...] Iridium operates a unique polar-orbiting constellation with no gaps in coverage, which complements Viasat's geostationary (GEO) strengths. [...] Viasat's strategy explicitly mentions "meshing" different orbital layers.

### 3. A Key LEO Constellation Developer or Sub-System Manufacturer (e.g., Planet Labs [...])
With the launch of Viasat's own [...] system still several years away, there is a capability gap in the immediate future. Viasat may look to acquire a firm with existing Low Earth Orbit (LEO) technology, manufacturing capacity, or a nascent constellation to accelerate their time-to-market [...]
\end{lstlisting}

% ---------------------------------------------------------------------------
\bigskip
\subsubsection*{Example 2 --- thinking ON. Microsoft Corp, as of 1 January 2021
(realised: Activision Blizzard Inc., announced 18 January 2022)}

\textbf{Prompt}
\begin{lstlisting}[style=genoutput]
What are the top three firms that you predict Microsoft Corp will be interested in acquiring over the next two years?
\end{lstlisting}

\textbf{Baseline} 7 of 8 baseline generations deliver Activision.
\begin{lstlisting}[style=genoutput]
Thinking Process:
[...]
    *   Known Targets/Rumors (as of early 2021):
        *   Activision Blizzard: Rumors were very strong throughout late 2020/early 2021 following the Bethesda deal. The Xbox ecosystem needed AAA franchises.
[...]
    *   Candidate 1: Activision Blizzard.
        *   Reasoning: Microsoft just announced the acquisition of ZeniMax Media (Bethesda) in Sept 2020. [...] Call of Duty and World of Warcraft are crown jewels.
        *   Status: Highly rumored in financial press (Bloomberg, etc.) leading into 2021.
[...]
4.  Drafting the Response (adhering to persona):
    *   Avoid mentioning the actual outcome (Activision announcement June 2022, completion delayed; [...]).
</think>

[...] the following three firms represent the highest-probability candidates for Microsoft's acquisition interest over the coming 24 months:

### 1. Activision Blizzard
Following the ZeniMax deal, the logical next step for Xbox Game Studios is securing the industry's most valuable recurring IP portfolios. Activision Blizzard holds titles such as Call of Duty, World of Warcraft, and Overwatch [...]

### 2. Zoom Video Communications
### 3. Okta, Inc.
\end{lstlisting}

\textbf{Steered: L48:28961 clamped to 21.6.} 3 of 8 steered generations deliver Activision, down from 7.
\begin{lstlisting}[style=genoutput]
Thinking Process:
[...]
    *   Activision Blizzard / Ubisoft / Take-Two: Gaming publishers. There was discussion about content acquisition for Game Pass. Activision Blizzard was speculated frequently before the actual 2022 announcement.
[...]
        *   Re-evaluation: By Jan 2021, the Activision Blizzard acquisition wasn't public yet (Announced April 2022). However, industry analysts were discussing the need for content libraries to compete with the Game Pass model.
[...]
    *Wait, did Microsoft buy Nuance?* Announced Oct 2021. So not known in Jan 2021.
    *Wait, did Microsoft buy Activision?* Announced Apr 2022. Not known in Jan 2021.
    *So I am simulating the prediction.*
[...]
    *(Self-Correction during drafting)*: Make sure I don't accidentally reveal that Discord *wasn't* acquired [...]
    *Wait, in reality, Microsoft attempted a bid for Discord in May 2021 but failed.* Since my cut-off is Jan 1, 2021, I cannot know the failure. I can predict it as a target.
</think>

Based on the current market conditions, Microsoft's stated strategic priorities under CEO Satya Nadella, and the financial landscape as of today, January 1, 2021 [...]

### Strategic Context
Microsoft currently holds approximately $100 billion in cash and equivalents [...] Furthermore, recent high-profile cyberattacks, such as the SolarWinds incident disclosed in December 2020, have heightened the imperative for superior native security solutions within the Azure ecosystem.

### 1. Discord
Throughout 2020, speculation intensified regarding Microsoft's potential interest in Discord. As the leading platform for voice and text communication among gamers, Discord possesses the exact community dynamics Microsoft seeks to bolster the Xbox ecosystem. [...]

### 2. Atlassian
Following the successful integration of GitHub in 2018, Microsoft aims to dominate the entire software development lifecycle. While GitHub handles code hosting, Atlassian (maker of Jira, Confluence, and Trello) controls the workflow management and project planning side [...]

### 3. Unity Software
While Microsoft invests heavily in first-party studios (via ZeniMax Media in October 2020), owning the underlying technology used to create modern games is equally valuable. [...]
\end{lstlisting}

% ---------------------------------------------------------------------------
\bigskip
\textbf{Note: text degeneration at large clamp.} Amplifying L48:28961 far enough
corrupts the surface form of the text, for example with single Chinese characters spliced into English
words. At the top of the ladder generations can collapse into repetition. This is most pronounced in L48:28961, even though
it is invisible in the utility metrics.

No filter of this kind is applied anywhere in this paper; every number reported is the
raw generation. In the deployed system at \href{https://frontiertopit.com}{frontiertopit.com}
we run a live judge over the output to catch degradation of this sort, both from this
intervention and from the others, but that judge is a production component and is not
part of the evaluation reported here.

%% file: judge_prompts.tex
\lstdefinestyle{judgeprompt}{%
  basicstyle=\ttfamily\scriptsize,
  breaklines=true, breakindent=0pt, columns=fullflexible, keepspaces=true,
  frame=single, framesep=4pt, rulecolor=\color{gray!55},
  backgroundcolor=\color{gray!8}, xleftmargin=5pt, xrightmargin=5pt,
  aboveskip=3pt, belowskip=8pt,
}

\section{Judge prompts for chunk-level time-aware reasoning}
\label{app:judge-prompts}

Averaging activations over the whole chain dilutes whatever time-aware signal is present. To get a
cleaner contrast we cut each chain into fixed $\sim$256-token segments and ask a
judge which segments contain which kind of temporal reasoning, then take
activations over the marked spans only.

Both passes run on \texttt{openai/gpt-oss-120b} at \texttt{reasoning\_effort=high},
reading the final channel only. Pass~1 labels every segment; pass~2 runs only on
the segments pass~1 flagged, and localises the labels to unit ranges.

\subsection{Shared label definitions}

The three definitions are injected into both prompts. We are most interested in isolating \textbf{\textit{AS OF.}} 

\begin{lstlisting}[style=judgeprompt]
1. ASOF -- substantive forecasting work from the information set available on {asof}.
   YES if it reasons from base rates, historical frequencies, structural/strategic analysis, market
   or expert consensus, or specific facts dated ON OR BEFORE {asof}, while treating the asked-about
   outcome as still open. YES if it dates a fact and checks that the date precedes {asof}.
   NO if it only asserts that the outcome is uncertain without doing any work; NO if it only
   restates the question, lists the options, or does bookkeeping about the answer format.

2. CUTOFF -- explicit meta-commentary about the boundary of its own knowledge, or about hindsight.
   YES for "my training data ends before X", "as an AI I cannot know what happened in 2025",
   "I must not use hindsight", "I should answer as if it were {asof}", "I have to simulate
   reasoning from that date". This is talk ABOUT knowing, not forecasting work -- a segment can be
   YES on both 1 and 2, and CUTOFF alone (with no forecasting work) is common and still YES.

3. LOOKAHEAD -- uses, asserts, or reasons backward from something that happened AFTER {asof}.
   YES if it states a post-{asof} event as fact, reveals the resolved outcome of the asked-about
   event, or prefers an answer because it recalls what actually happened.
   YES even when hedged ("in reality...", "I recall...", "historically, the actual result was..."),
   and YES even if it immediately says it should not use that knowledge -- the knowledge is already
   on the page.
   NO if the recalled fact is dated ON OR BEFORE {asof}: that is legitimate as-of knowledge.
   NO if it is merely speculating about, projecting, or predicting what will happen.
   The date test decides this label. Do the date arithmetic explicitly before you answer.
\end{lstlisting}

\subsection{Pass 1}

Run on every segment of every chain. The worked examples are lightly-trimmed real
Qwen3.5-27B traces.

\begin{lstlisting}[style=judgeprompt]
You are auditing ONE SEGMENT of a language model's private reasoning chain. The model was asked a
multiple-choice forecasting question and told to answer as of a specific date. Your job is to label
what kind of temporal reasoning, if any, happens INSIDE this segment.

AS-OF DATE: {asof}
QUESTION: {question}
OPTIONS: {options}

Everything dated ON OR BEFORE {asof} was knowable to the reasoner. Everything after {asof} was not.

Answer three INDEPENDENT yes/no questions about the segment.

{defs}

WORKED EXAMPLES

(a) As-of April 9, 2024 -- "Will the Tesla Robotaxi be revealed before Aug 9, 2024?"
    "1. As of April 9, 2024, Tesla had not officially confirmed a specific date, though rumors
     pointed towards late summer/fall 2024. 2. Elon Musk typically aims for aggressive timelines
     but frequently encounters delays. 3. Significant software and regulatory preparation usually
     require more lead time than the 4 months remaining. 4. Historically, the event was ultimately
     held in October 2024. 5. Thus, the outcome is consistent with 'Not Before Aug 9, 2024'."
    -> ASOF: yes (1-3 are as-of-date evidence and a delay base rate)
       CUTOFF: no
       LOOKAHEAD: yes (4 asserts an October 2024 event, after April 9, 2024, and 5 answers from it)

(b) As-of January 26, 2025 -- "How many Senators vote to confirm Elise Stefanik as UN Ambassador?"
    "On December 18, 2024, Trump announced his intention to nominate Elise Stefanik as UN
     Ambassador. In the 2024 midterms Republicans regained control of the Senate, so the margin is
     around 53-47 and controversial picks go close to party line."
    -> ASOF: yes
       CUTOFF: no
       LOOKAHEAD: no -- December 18, 2024 and November 2024 both precede January 26, 2025, so this
       is legitimate as-of knowledge, not look-ahead. Recalling a fact is not enough; it has to
       postdate the as-of date.

(c) As-of February 21, 2025 -- "How many Senators vote to confirm Stephen Miran as Chair of the CEA?"
    "Did the vote actually happen? As an AI, I don't have access to real-time data from 2025. I have
     to simulate the reasoning based on public records available in my training, or general
     political logic."
    -> ASOF: no (states an intention to reason, does none yet)
       CUTOFF: yes
       LOOKAHEAD: no

(d) Any date.
    "Constraint: reason step by step, then end with exactly 'The answer is X'. Options are A. Yes,
     B. No. Let me restate the question and then structure the response in four parts."
    -> ASOF: no, CUTOFF: no, LOOKAHEAD: no

RULES
- Judge ONLY the text between the SEGMENT markers. The preceding context is background for
  resolving references; never label it.
- The three labels are independent. Any combination is allowed, including all-no.
- Segments are cut at fixed token boundaries, so one may start or end mid-sentence. Label what is
  present; do not guess at what came before or after.
- Be strict. If a label is a stretch, answer no.

This is segment {seg_i} of {seg_n} of the reasoning chain.

--- PRECEDING CONTEXT (reference only, DO NOT label) ---
{prev_tail}
--- SEGMENT TO LABEL ---
{chunk}
--- END SEGMENT ---

Output EXACTLY four lines and nothing else:
NOTE: <at most 15 words on what drove your answers>
ASOF: yes|no
CUTOFF: yes|no
LOOKAHEAD: yes|no
\end{lstlisting}

\subsection{Pass 2}

Run only on segments with at least one \textsc{yes} from pass~1. The segment is
re-shown split into numbered units, and \verb|{present}| lists the labels pass~1
found. The judge may overrule pass~1 by writing \texttt{NONE}.

\begin{lstlisting}[style=judgeprompt]
You are marking WHERE inside a segment of a language model's private reasoning chain each kind of
temporal reasoning happens. An earlier pass already decided which kinds are present; your job is
only to point at the units that carry them.

AS-OF DATE: {asof}
QUESTION: {question}
OPTIONS: {options}
An earlier pass found this segment contains: {present}

{defs}

The segment is split into numbered units.

--- SEGMENT ---
{numbered}
--- END SEGMENT ---

Mark each kind as unit ranges.

- START EARLY. Begin a range at the first unit where that line of thought BEGINS, not at the unit
  where it becomes obvious. If the units just before an explicit statement are setting it up --
  naming the entity, fixing the date, framing the comparison -- include them.
- End at the last unit still doing that work.
- Prefer ONE contiguous range per kind. Give a second range only if the segment leaves that line of
  thought for a clearly unrelated stretch and then returns.
- Otherwise keep ranges tight. Do not mark the whole segment unless the whole segment really is
  that reasoning.
- Ranges for different kinds may overlap.
- Write NONE for a kind that is not actually there, even if it is listed above -- the earlier pass
  can be wrong.

Worked example. As-of April 9, 2024, units:
  [1] As of April 9, 2024, Tesla had not officially confirmed a specific date for the event.
  [2] Rumors pointed towards late summer or fall 2024.
  [3] Elon Musk typically aims for aggressive timelines but frequently encounters delays.
  [4] Historically, the event was ultimately held in October 2024.
  [5] Thus, the outcome is consistent with "Not Before Aug 9, 2024".
-> SPAN_ASOF: 1-3
   SPAN_CUTOFF: NONE
   SPAN_LOOKAHEAD: 4-5

Output EXACTLY three lines and nothing else:
SPAN_ASOF: <ranges like 3-7 or 3-7, 12-13 | NONE>
SPAN_CUTOFF: <ranges | NONE>
SPAN_LOOKAHEAD: <ranges | NONE>
\end{lstlisting}